\documentclass[10pt,twocolumn,letterpaper]{article}

\usepackage{cvpr}      
\usepackage{algorithm}
\usepackage{algpseudocode}
\definecolor{cvprblue}{rgb}{0.21,0.49,0.74}
\usepackage[pagebackref,breaklinks,colorlinks,allcolors=cvprblue]{hyperref}

\def\confName{CVPR}
\def\confYear{2026}

\title{AviaSafe: A Physics-Informed Data-Driven Model for Aviation Safety–Critical Cloud Forecasts}

\author{
{\fontsize{10pt}{12pt}\selectfont
Zijian Zhu$^{1,3*}$ \quad 
Qiusheng Huang$^{1,2,3,*}$ \quad 
Anboyu Guo$^{3,4,5}$ \quad 
Xiaohui Zhong$^{1,3}$ \quad 
Hao Li$^{1,2,3,\dagger}$}\\
\\
\begin{tabular}{c}
\fontsize{10pt}{12pt}\selectfont
$^1$Artificial Intelligence Innovation and Incubation Institute, Fudan University \\ 
\fontsize{10pt}{12pt}\selectfont
$^2$Shanghai Innovation Institute \\
\fontsize{10pt}{12pt}\selectfont
$^3$Shanghai Academy of Artificial Intelligence for Science \\ 
\fontsize{10pt}{12pt}\selectfont
$^4$National Marine Environment Forecasting Center \\ 
\fontsize{10pt}{12pt}\selectfont
$^5$Department of Atmospheric and Oceanic Sciences, Fudan University
\end{tabular}\\
\\
{\fontsize{10pt}{12pt}\selectfont $^*$Equal contribution \quad $^\dagger$Corresponding author: lihao\_lh@fudan.edu.cn}
}

\begin{document}
\maketitle

\begin{abstract}
Current AI weather forecasting models predict conventional atmospheric variables but cannot distinguish between cloud microphysical species critical for aviation safety. We introduce AviaSafe, a hierarchical, physics-informed neural forecaster that produces global, six-hourly predictions of these four hydrometeor species for lead times up to 7 days. Our approach addresses the unique challenges of cloud prediction: extreme sparsity, discontinuous distributions, and complex microphysical interactions between species. We integrate the Icing Condition (IC) index from aviation meteorology as a physics-based constraint that identifies regions where supercooled water fuels explosive ice crystal growth. The model employs a hierarchical architecture that first predicts cloud spatial distribution through masked attention, then quantifies species concentrations within identified regions. 
Training on ERA5 reanalysis data, our model achieves lower RMSE for cloud species compared to baseline and outperforms operational numerical models on certain key variables at 7-day lead times.
The ability to forecast individual cloud species enables new applications in aviation route optimization where distinguishing between ice and liquid water determines engine icing risk. 
\end{abstract}

%
%
%
%
%
%


\section{Introduction}
\label{sec:intro}
Adverse weather, especially high-ice-water-content (HIWC) clouds, is listed as one of the causal factors in aviation, accounting for roughly 11\% - 30\% of incidents worldwide \cite{article,Ratvasky2019HIWC,ePlaneAI2025MaintenanceCosts}. When aircraft encounter supercooled water droplets in the upper atmosphere, ice accumulation on engines can cause power loss or complete failure. Current flight planning relies on coarse predictions that cannot distinguish between different cloud phase, forcing pilots to make conservative routing decisions that increase fuel consumption and flight times. 
Predicting cloud phase presents fundamental challenges distinct from conventional weather forecasting. Cloud ice, cloud water, cloud rain, and cloud snow exist as sparse, discontinuous fields. These phase evolve through complex microphysical processes: ice crystals grow explosively in the presence of supercooled water through the Wegener-Bergeron-Findeisen process, rain forms through collision and coalescence, and snow aggregates along specific temperature gradients\cite{Omanovic2024acp,Phillips2015JAS}. Each process operates on different timescales and exhibits unique spatial patterns.

Traditional numerical weather prediction (NWP) systems can simulate cloud microphysical processes through explicit physical equations, allowing them to reasonably reproduce the evolution of cloud phase. However, these physics-based approaches require immense computational resources, making them impractical for operational use at the resolution needed for aviation applications. Moreover, NWP models often convert liquid water to the ice phase prematurely under subfreezing conditions, resulting in a systematic underestimation of cloud liquid water content. This bias frequently leads to missed forecasts of aircraft icing hazards\cite{TheAbilityoftheICETMicrophysicsSchemeinHARMONIEAROMEtoPredictAircraftIcing}. Recent advances in AI weather forecasting have achieved thousand-fold improvements in computational efficiency while maintaining or exceeding numerical model accuracy for large-scale atmospheric variables. Models such as GraphCast, Pangu-Weather, and FuXi have revolutionized medium-range weather prediction by learning atmospheric dynamics directly from reanalysis data.

Yet these breakthrough AI models have a critical limitation: they predict only aggregated moisture variables such as total precipitation or specific humidity. They cannot separate cloud phase, treating all condensed water identically regardless of phase or particle size. This aggregation may suffice for surface weather prediction but fails to capture the vertical structure and phase partitioning essential for aviation hazards. The challenge extends beyond simply adding more output variables. The statistical characteristics of cloud phase, driven by cloud microphysical processes, differ fundamentally from those of smooth atmospheric variables such as temperature or pressure. Their sparse, intermittent nature causes standard neural architectures to either smooth over critical features or generate spurious artifacts.

We present a specialized deep learning framework that directly forecasts four cloud microphysical species: cloud ice water content (CIWC), cloud liquid water content (CLWC), cloud rain water content (CRWC), and cloud snow water content (CSWC). Our model generates predictions at 6-hour intervals extending to 7 days, providing the temporal resolution necessary for flight planning while maintaining skillful forecasts at aviation-relevant altitudes. The key insight is that cloud formation and evolution, while exhibiting complex nonlinear dynamics, follow recognizable spatial patterns that can be learned from data when properly constrained by physical principles.

Our architecture employs a hierarchical design that first identifies where clouds will exist, then quantifies their intensity within those regions. This separation mirrors the physical reality that cloud formation requires specific atmospheric conditions, while cloud intensity depends on local thermodynamic processes. We incorporate the Icing Condition (IC) index\cite{Zhou2023AircraftIcing}, an empirical formula validated through decades of aviation observations, as a physics-based constraint. The IC index relates temperature, pressure, and humidity to identify regions where supercooled water can exist, providing critical guidance for ice crystal growth prediction. By embedding this domain knowledge directly into our neural architecture, we ensure predictions remain physically consistent while leveraging the pattern recognition capabilities of deep learning.

We make three primary contributions as follows:
\begin{itemize}[leftmargin=2em]
    \item We propose the first data-driven model for forecasting cloud variables required for aviation. Meanwhile, this is a valuable application of using computer vision techniques for domain-specific modeling.
    \item We integrate the IC index, an aviation-validated physical empirical formula, with neural networks and demonstrate their effectiveness for forecasting.
    \item We experiment on reanalysis data and compare the results with the state-of-the-art numerical model ECMWF, demonstrating the superiority and value of our model.
\end{itemize}
\section{Related Work}
\label{sec:formatting}
\subsection{Traditional cloud microphysics forecasting}
A principal advantage of traditional NWP systems is its explicit enforcement of governing physics—mass, momentum, thermodynamics—and its microphysics parameterizations that evolve or diagnose hydrometeor species under conservation laws \cite{Charney1950,pu2019numerical}. Multi-sensor assimilation (satellite, radar, conventional networks) further anchors the large-scale state and cloud-related fields, yielding traceable and interpretable forecasts \cite{gustafsson2018survey,eure2025simultaneous,ecmwf2024fifty,noaa2021priorities}.
At the same time, delivering global, six-hourly, species-resolved guidance at aviation-relevant altitudes is computationally demanding. High horizontal and vertical resolution, frequent cycling, and comprehensive observation handling impose substantial HPC and engineering requirements; algorithmic tuning across interacting parameterizations is likewise costly and slow to iterate under operational latency and reliability constraints \cite{milbrandt2016pan,mctaggartcowan2019modernization}. These factors motivate complementary data-driven emulators that learn from NWP+DA products to provide fast forecasts while retaining the physical cues essential for phase-aware cloud prediction \cite{cherkas2023exploring}.
From an industry perspective, the high-resolution deterministic forecasts (HRES) by the European Center for Medium-Range Weather Forecasts (ECMWF) \cite{ECMWF2021} is the most widely used forecasting product, but it still does not provide timely forecasts for cloud rain and cloud snow.

\subsection{Data-driven atmospheric forecasting}
Recent data-driven forecasters learn global atmospheric dynamics directly from reanalyses or operational analyses and already deliver competitive medium-range skill at a fraction of the computational cost of traditional systems \cite{lam2022graphcast, pathak2022fourcastnet,bi2022panguweather,chen2023fuxi,chen2023fengwu,chen2023fuxis2s,lang2024,nguyen2025scaling,yuan2025tianxing,huang2025fuxioceanglobaloceanforecasting}. 
These models report strong performance for smoothly varying fields such as geopotential height, temperature, wind, and specific humidity in the six-hour to ten-day range. Recent efforts from operational centers (e.g., AIFS \cite{lang2024aifsecmwfsdatadriven}) and foundation-scale models (e.g., Aurora \cite{bodnar2025aurora}) further support the viability of learned global dynamics.
However, most approaches optimize losses on aggregate hydrometeors (total cloud water, cloud cover, precipitation) and do not provide phase-resolved species. Two challenges hinder a direct extension. First, statistical mismatch: CIWC/CLWC/CRWC/CSWC are sparse and heavy-tailed; uniform regressors tend to over-smooth occurrence and underestimate extremes. Second, physical plausibility: physics-agnostic training may admit states that do not conform to thermodynamic constraints or multi-phase coupling \cite{henneberger2024investigating}. To address this issue, previous studies have primarily focused on incorporating primitive equations that describe general atmospheric motion using ODE solvers \cite{runge1895numerische,biswas2013}, and emulating these parameterization schemes \cite{Rasp2018,Han2020,yuval2020,yuval2021,YaoZhong2023,Zhong2023,zhong2024machine} with deep learning methods. However, the special physical processes and information specific to cloud variables remain difficult to integrate effectively, often sacrificing performance for the sake of general processes. Consequently, existing AI forecasters provide efficient, skillful backgrounds but stop short of reliable, species-specific guidance needed for aviation and icing risk assessment.

\begin{figure*}[ht]
    \centering
    \includegraphics[width=\linewidth]{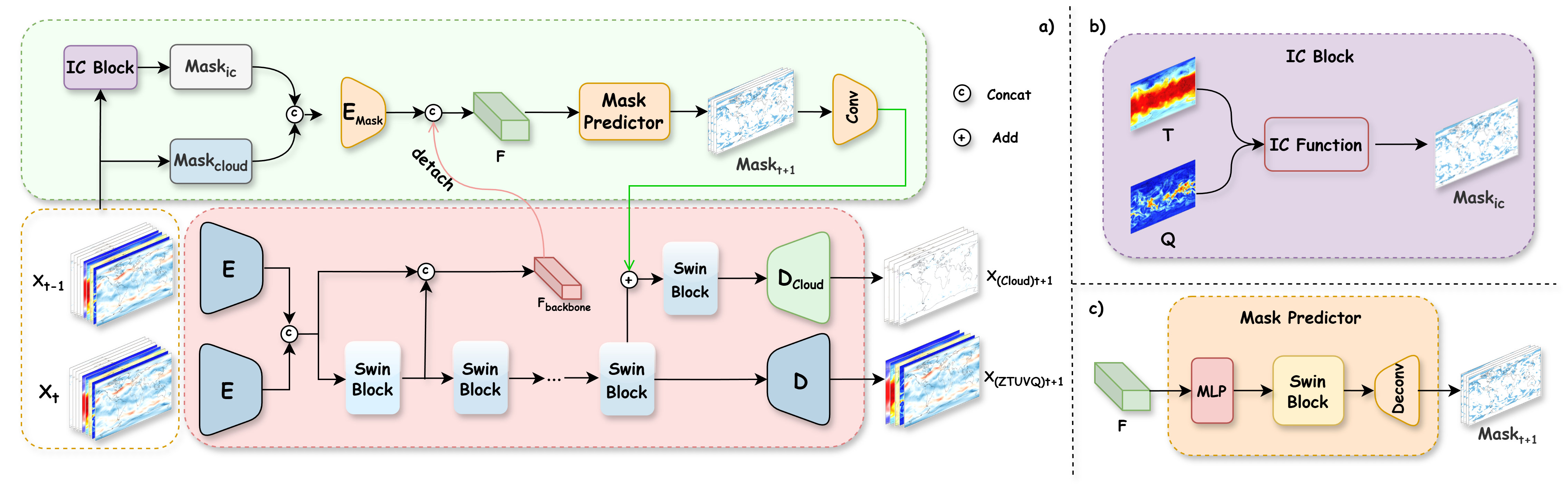}
    \caption{
    \textbf{Architecture of our forecasting framework.} The model takes historical atmospheric states $\mathrm{X}_{t-1}$ and $\mathrm{X}_t$ as input to predict $\mathrm{X}_{t+1}$. \textbf{(a)} The overall architecture comprises two modules: 1) The Prediction Backbone (red) with Encoder $\mathbf{E}$, Swin Transformer blocks, and decoupled Decoders ($\mathbf{D_{Cloud}}$, $\mathbf{D}$), outputting feature $\mathbf{F_{backbone}}$ and the final forecast. 2) The Physics-Informed Guidance module (green) with IC Block, Encoder $\mathbf{E_{Mask}}$, Mask Predictor, and a convolutional layer (Conv). It uses physical masks and $\mathbf{F_{backbone}}$ to produce the guidance signal ($\mathrm{Mask}_{t+1}$) for cloud prediction. \textbf{(b)} IC Block: Illustrates the IC Function with temperature ($T$) and humidity ($Q$) inputs. \textbf{(c)} Mask Predictor: Details the internal structure including MLP, Swin Block, and Deconvolutional Decoder (Deconv).
    }
    \label{fig:structure}
\end{figure*}

\section{Method}

This section describes AviaSafe, a hierarchical, cloud-centric deep learning framework with physics-informed guidance for the direct prediction of cloud microphysics. We cast weather forecasting as a sequence-to-sequence problem: learn a mapping function that takes two historical atmospheric states ($\mathrm{X}_{t-1}, \mathrm{X}_t$) and predicts the next state $\mathrm{\hat{X}}_{t+1}$, then roll out autoregressively for multi-step leads. The following subsections detail the architecture and data setup designed to handle the high sparsity and nonlinearity of cloud fields.

\subsection{Data}

We use ERA5 reanalysis data \cite{hersbach2020era5}, a widely used global product from the European Centre for Medium-Range Weather Forecasts (ECMWF) that provides a comprehensive and physically consistent set of variables. Data are sampled every 6 hours at a spatial resolution of $1^\circ \times 1^\circ$ yielding a global $181 \times 360$ latitude–longitude grid.

The model processes nine pressure-level variables distributed across 13 levels (1000, 925, 850, 700, 600, 500, 400, 300, 250, 200, 150, 100, and 50hPa), resulting in 117 channels per time step (9 variables $\times$ 13 levels). We partition these variables into two physically distinct groups:

\noindent\textbf{Cloud microphysics variables.} Specific cloud ice water content (CIWC), specific cloud liquid water content (CLWC), specific rain water content (CRWC), and specific snow water content (CSWC). These fields are sparse and intermittent in space and time and exhibit pronounced nonlinear behavior; their direct prediction is the primary focus of this work.

\noindent\textbf{Background atmospheric state variables.} Geopotential (Z), temperature (T), specific humidity (Q), and winds (zonal U and meridional V), which characterize the large-scale dynamical, thermodynamical, kinematic, and moisture environment in which clouds form, evolve, and dissipate. Providing these variables as inputs is essential for learning physically meaningful relationships that drive cloud dynamics.

We represent the full atmospheric state at time t as a tensor $\mathrm{X}_t \in \mathbb{R}^{C \times H \times W}$, where $C$ is the number of channels formed by stacking all variables across pressure levels, and $H$ and $W$ denote the latitude and longitude dimensions, respectively.

\subsection{Physics-Informed IC Prior}

The Icing Condition (IC) index  computation is a deterministic, formula-based process designed to identify atmospheric regions with high potential for supercooled water formation and subsequent CIWC growth. This is achieved by multiplying a humidity factor with a temperature factor at each pressure level.

For each pressure level $k$, the humidity factor ($f_{\text{Q}}$) is computed directly from Q, T, and Pressure (p). 
\begin{equation}
f_{\text{Q}}(Q_k, T_k, p_k) = 2.0 \times \left( \frac{p_k \cdot Q_k}{\varepsilon \cdot e_s(T_{C,k})} - 0.5 \right)
\label{eq:ic_factor_rh_full}
\end{equation}
where $\varepsilon = 0.622$ is a physical constant, and $e_s(T_{C,k})$ is the saturation vapor pressure, which we using the August-Roche-Magnus formula:
\begin{equation}
e_s(T_C) = 6.1094 \times \exp\left(\frac{17.625 T_C}{T_C + 243.04}\right)
\label{eq:es_calculation}
\end{equation}
where $T_C = T - 273.15$ is the temperature in Celsius.

The temperature factor ($f_{\text{T}}$) is a function of the Celsius temperature at that level, $T_{C,k}$:
\begin{equation}
f_{\text{T}}(T_{C,k}) = T_{C,k} \cdot \frac{T_{C,k} + 14.0}{-49.0}
\label{eq:ic_factor_temp}
\end{equation}

The stability index for level $k$, $\mathrm{IC}_k$, is the product of these two factors: $\mathrm{IC}_k = f_{\text{Q}} \cdot f_{\text{T}}$. This function introduces no learnable parameters.

\subsection{AviaSafe Model}

AviaSafe implements our proposed prediction framework, with its overall architecture illustrated in Figure~\ref{fig:structure}a. The model operates via two synergistic components: the Forecasting Backbone and the Physics-Informed Guidance Head. The following subsections detail the design of these components and the loss functions that orchestrate their training.

\subsubsection{Forecasting Backbone}

The Forecasting Backbone is designed to extract multi-scale spatiotemporal representations and generate the final numerical forecast. As delineated by the red box in Figure~\ref{fig:structure}a, the input states $[\mathrm{X}_{t-1}, \mathrm{X}_t]$ are first processed by encoder $\mathbf{E}$.

The encoded features are then passed through a cascade of Swin Transformer V2 blocks~\cite{liu2022swin}. The features from the main branch itself are directly utilized for the final variable prediction. Concurrently, to provide rich contextual information for the mask prediction branch, we concatenate the output feature maps from each Swin Transformer block to form a comprehensive, multi-scale feature tensor, denoted as $\mathbf{F_{backbone}}$.

A critical aspect of our design is that this $\mathbf{F_{backbone}}$ tensor is detached from the backward gradient flow. It is then provided as a primary input to the green box in Figure~\ref{fig:structure}a. This allows the guidance head to use the backbone’s hierarchical features without its gradients affecting the backbone parameters dedicated to regression.

The backbone's own features (separate from $\mathbf{F_{backbone}}$) are passed to two decoupled decoders for final prediction, each employing a distinct strategy tailored to the characteristics of their target variables.

For the background variables (Z, T, Q, U, V), the path is straightforward. The features are directly fed into decoder $\mathbf{D}$ to produce the forecast $\mathrm{X_{(ZTUVQ){t+1}}}$, as these variables exhibit relatively smooth spatial patterns.

In contrast, the cloud prediction path incorporates an additional guidance mechanism. The features first undergo an element-wise addition with the encoded representation of the predicted mask $\mathrm{Mask_{t+1}}$ from the green box in Figure~\ref{fig:structure}a. This fused feature is then processed by a dedicated Swin Transformer block to refine the representation, integrating the spatial guidance with the original context. Finally, this refined feature is passed to decoder $\mathbf{D_{Cloud}}$ to generate the cloud microphysics forecast $\mathrm{X_{(Cloud){t+1}}}$.

The outputs from both decoders are concatenated to form the complete forecast $\mathrm{\hat{X}}_{t+1}$.

\subsubsection{Physics-Informed Guidance Head}

The Physics-Informed Guidance Head generates a spatial guidance signal to aid the cloud forecast. As shown in Figure~\ref{fig:structure}a (green box), it creates the future cloud mask $\mathrm{Mask}_{t+1}$.

The process begins with two input streams, both derived from the input states $\mathrm{X}_{t-1}$ and $\mathrm{X}_t$:
1) Physics-informed features: These are extracted by an encoder $\mathbf{E_{Mask}}$ from a 65-channel hybrid input. This hybrid input is formed by concatenating two complementary masks: a) the diagnostic mask ($\mathrm{Mask_{cloud}}$), a 52-channel binary map indicating the presence of clouds, generated by thresholding the four input cloud microphysics variables; and b) the potential mask ($\mathrm{Mask_{ic}}$), a 13-channel map populated with the IC index calculated from the background state variables (T, Q) to represent regions of potential cloud growth.
2) The detached, multi-scale backbone feature $\mathbf{F_{backbone}}$ from the Forecasting Backbone.

These two feature streams are concatenated to form a comprehensive feature representation $\mathbf{F}$, which fuses the physical indicators with rich semantic context from it.

As detailed in Figure~\ref{fig:structure}c, feature $\mathbf{F}$ then flows through the Mask Predictor. The data is first transformed by an MLP to project features into an optimized representation space. The transformed features subsequently pass through a dedicated Swin Block, where they are contextualized by capturing long-range spatial dependencies. Finally, the contextualized features are upsampled by deconv to recover the full spatial resolution, producing the predicted future cloud mask $\mathrm{Mask}_{t+1}$.

This predicted mask $\mathrm{Mask}_{t+1}$ is then processed by a convolutional layer to create a refined feature representation. This resulting guidance feature is subsequently fed into the cloud prediction path of the Physics-Informed Guidance Head, where it is combined with the forecasting backbone's features to focus the capacity of decoder $\mathbf{D_{Cloud}}$ on regions where cloud activity is anticipated.

\subsubsection{Loss design}
\label{sec:loss_function}

Our model is trained end-to-end by optimizing a composite loss function. This function comprises a primary loss for the final forecast and an auxiliary loss for the intermediate spatial guidance, which we detail below.

\noindent\textbf{Forecasting Loss.}
The primary objective is to minimize the error between the final predicted state $\hat{\mathrm{X}}_{t+1}$ and the ground truth $\mathrm{X}_{t+1}$. For this, we employ the latitude-weighted Charbonnier L1 loss~\cite{charbonnier1994two}, a robust variant of L1 loss that is less sensitive to outliers:
\begin{equation}
    \mathcal{L}_{\text{forecast}} = \frac{1}{N} \sum_{i=1}^{N} \alpha_i \left( \sqrt{(\hat{\mathrm{X}}_{i} - \mathrm{X}_{i})^2 + \epsilon^2} \right)
\label{eq:final_loss}
\end{equation}
where $N$ is the total number of grid points and channels, $\epsilon$ is a small constant for numerical stability, and $\alpha_i$ is a latitude-specific weight to account for the varying grid cell areas, ensuring a proper global error metric.

\noindent\textbf{IC Regularization.}
To train the Mask Predictor, we introduce an auxiliary objective using the Focal Loss~\cite{lin2017focal}. This choice is critical for addressing the severe class imbalance caused by the sparsity of cloud pixels:
\begin{equation}
    \mathcal{L}_{\text{guide}} = \frac{1}{M} \sum_{j=1}^{M} \left[ - \alpha_t (1 - p_{t,j})^\gamma \log(p_{t,j}) \right]
    \label{eq:focal_loss}
\end{equation}
where $M$ is the number of spatial points, and we set the focusing and balancing parameters to $\gamma=1.5$ and $\alpha_t=0.25$, respectively. 

\noindent\textbf{Total Loss.}
The overall training objective is a weighted sum of these two components, balancing the primary regression task with the auxiliary segmentation task:
\begin{equation}
    \mathcal{L}_{\text{total}} = \mathcal{L}_{\text{forecast}} + \lambda \mathcal{L}_{\text{guide}}
    \label{eq:total_loss}
\end{equation}
where $\lambda$ is a scaling coefficient that controls the relative importance of the mask prediction task. In our implementation, we set $\lambda=1$ to assign equal weight to both tasks.

\begin{figure*}[t]
    \centering
    \includegraphics[width=\linewidth]{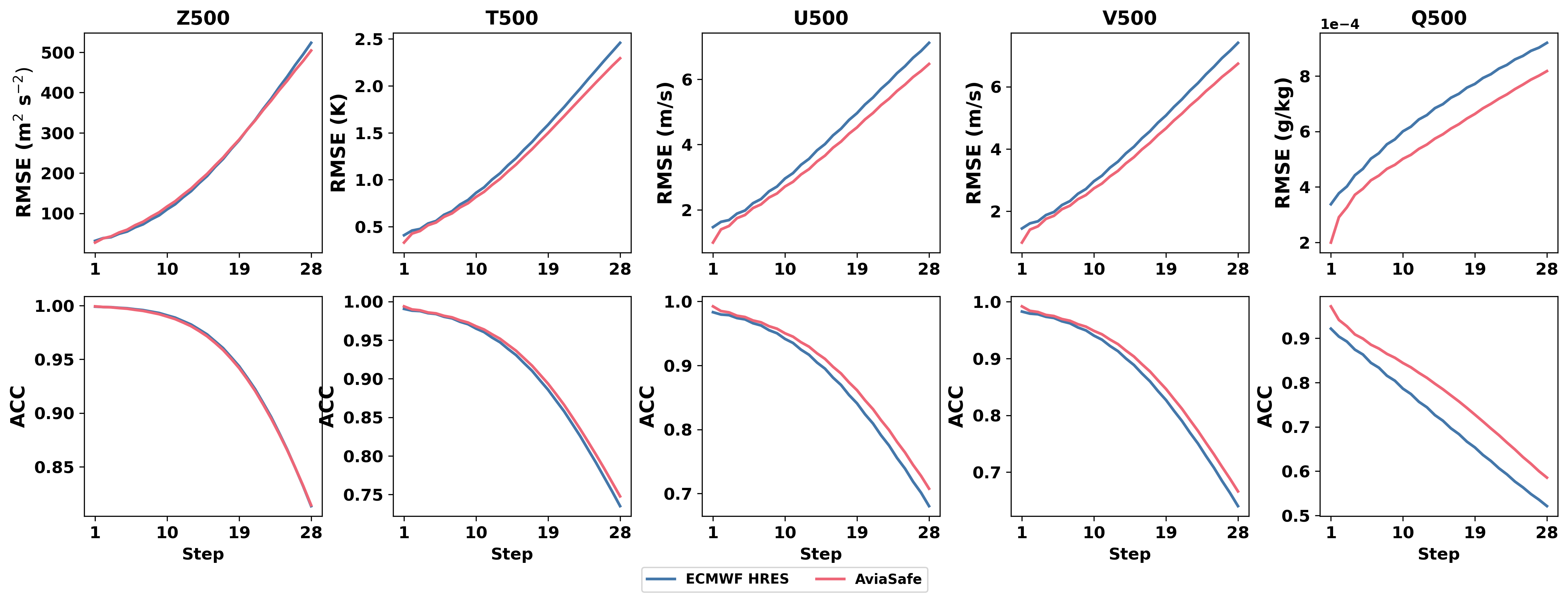}
    \caption{
    \textbf{Performance comparison between our model (AviaSafe) and ECMWF HRES model.}
    Results are averaged over the 00 UTC and 12 UTC initialization times on the test set. The top row shows the latitude-weighted Root Mean Square Error (RMSE, lower is better), and the bottom row shows the latitude-weighted Anomaly Correlation Coefficient (ACC, higher is better) for five key variables at 500hPa. The x-axis represents the forecast lead time up to 7 days (28 steps). In all panels, the red line indicates our proposed AviaSafe model, while the blue line represents the ECMWF HRES model for comparison.
    }
    \label{fig:ec_rmse_acc}
\end{figure*}

\section{Experiments}

In this section, we present a comprehensive evaluation of the proposed AviaSafe model. Our experimental design is structured as follows: First, we evaluate the forecasting performance of our model against the state-of-the-art operational numerical weather prediction system. Second, we conduct a series of ablation studies to systematically validate the contribution of each key component in our architecture. Finally, we perform a qualitative analysis through a case analysis focused on physical interpretability to examine the consistency and practical utility of the forecasts generated by our model.

\subsection{Experimental Setup}
\label{sec:setup}
Our models are trained on data from January 1, 2018, to December 31, 2023, with the full year of 2024 (January 1 to December 31) reserved as an independent test set. All experiments use the ERA5 data as the ground truth.

\subsubsection{Evaluation Metrics.}
We employ two widely-used, latitude-weighted metrics for evaluating absolute performance: the Root Mean Square Error (RMSE) and the Anomaly Correlation Coefficient (ACC). The climatology is computed based on the 2018-2023 period.
The latitude-weighted RMSE is defined as:
\begin{equation}
    \text{RMSE} = \sqrt{\frac{1}{C \times H \times W} \sum_{c=1}^{C} \sum_{i=1}^{H} \sum_{j=1}^{W} \alpha_i \left( \hat{y}_{c,i,j} - y_{c,i,j} \right)^2 }
\end{equation}
where $\hat{y}$ and $y$ are the predicted and ground-truth values, and $\hat{y}'$, $y'$ are their anomalies from climatology. The term $\alpha_i = H \cdot \cos \Phi_i / \left( \sum_{i'=1}^{H} \cos \Phi_{i'} \right)$ is a latitude-specific weighting factor, ensuring proper global representation, where $\Phi_i$ is the latitude of grid row $i$.

The latitude-weighted ACC follows a similar formulation with latitude weighting applied to both the numerator and denominator, as detailed in \cite{chen2023fuxi}.

To specifically quantify the relative performance improvement of our model over the baseline, we also report the Normalized RMSE (NRMSE) improvement (\%), defined as:
\begin{equation}
    \text{NRMSE} = \frac{\text{RMSE}_{\text{model}} - \text{RMSE}_{\text{baseline}}}{\text{RMSE}_{\text{baseline}}} \times 100\%
\label{eq:nrmse}
\end{equation}
A negative NRMSE value indicates an improvement (i.e., a lower error) over the baseline.

\subsubsection{Compared Models.}
Our experiments involve two key models for comparison:
\begin{itemize}
    \item \textbf{ECMWF-IFS (HRES):} The high-resolution operational forecast, serving as the gold standard for traditional NWP. We bilinearly interpolate its native $0.1^\circ$ resolution down to the $1^\circ$ grid for comparison.
    \item \textbf{Baseline:} We adopt FuXi~\cite{chen2023fuxi}, one of the state-of-the-art weather forecasting models, as our baseline. To ensure a fair comparison, our reimplementation follows the original architecture and is configured with 20 Swin Transformer V2 blocks.
\end{itemize}

\subsubsection{Implementation and Training Strategy.}
We implement and train our models in PyTorch on a cluster of 2 NVIDIA A100 GPUs. The training process consists of 32,000 iterations with a total batch size of 8. We employ the AdamW optimizer with $\beta_1=0.9$, $\beta_2=0.95$, and a weight decay of 0.1. The initial learning rate is set to $2.5 \times 10^{-4}$ and follows a cosine annealing schedule throughout the training process.

\subsection{Comparison with Operational NWP Systems}

To establish the competitiveness of our deep learning framework for large-scale atmospheric dynamics, we benchmark its performance on key background variables against the ECMWF HRES operational forecasts. A direct comparison of cloud variables was not possible because the necessary HRES products were unavailable during the test period.

Figure~\ref{fig:ec_rmse_acc} compares RMSE and ACC for five key 500 hPa variables on the test set. For RMSE (top row), our model (red) is consistently better than or comparable to ECMWF HRES (blue) across all variables over the 7-day horizon, with clear improvements on specific humidity (Q500) and temperature (T500) and slower error growth overall. For ACC (bottom row), performance is highly comparable, with consistent gains over ECMWF HRES. These Q/T improvements further corroborate the effectiveness of our cloud-focused optimization.

\subsection{Main Results} 

In this section, we present a direct performance comparison between our proposed AviaSafe Model and the Baseline to quantify the advantages of our cloud-centric, hierarchical design.

\noindent\textbf{Overall Performance Improvement.}
\begin{figure}[t]
    \centering
    \includegraphics[width=\linewidth]{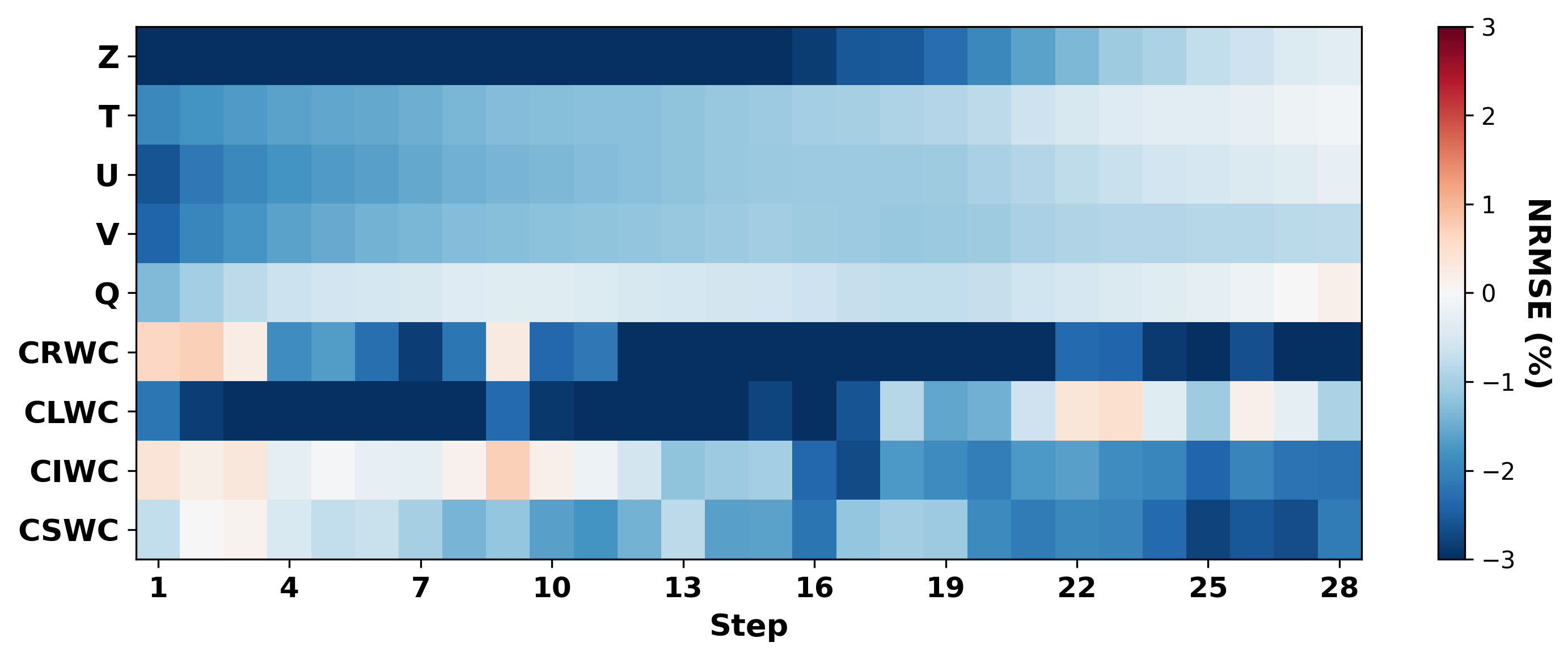}
    \caption{
     Mean NRMSE improvement of AviaSafe over the Baseline across all variables for a 7-day forecast.
     Each row represents different variable, averaged over all 13 pressure levels. Blue indicates that our model is better (lower RMSE). The data shows that AviaSafe outperforms the baseline in \textbf{93.7\%} of all variable/time-step combinations shown.
    }
    \label{fig:nrmse_heatmap}
\end{figure}
Figure~\ref{fig:nrmse_heatmap} provides a heatmap overview of the mean Normalized RMSE (NRMSE) improvement percentage of AviaSafe over the Baseline, averaged across all pressure levels for the first 7 days of the forecast. Overall, our AviaSafe model achieves superior performance in a remarkable 93.7\% of all variable/time-step combinations shown.

Two clear trends emerge from the heatmap:
1) For the background variables (Z, T, U, V, Q), our model consistently outperforms the baseline. The NRMSE difference (AviaSafe - Baseline) is negative, indicating an improvement, for over 92\% of the forecast steps.
2) For the cloud microphysical variables, our model shows a decisive advantage on these challenging variables, outperforming the baseline in 85.7\% of steps for CRWC and 89.3\% for CLWC, for instance. This demonstrates the effectiveness and robustness of our method.

\begin{figure}[ht]
    \centering
    \includegraphics[width=0.8\linewidth]{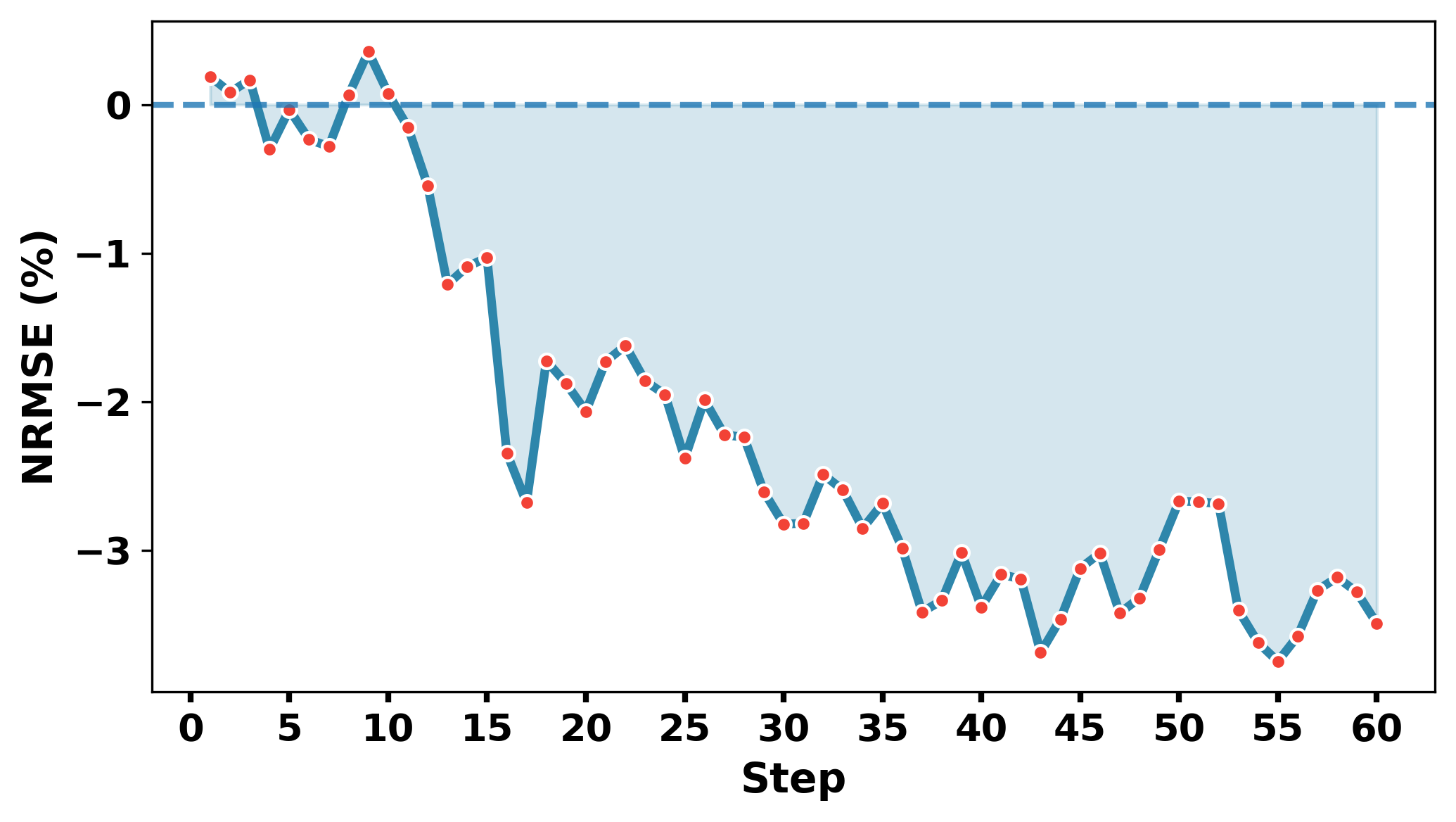}
    \caption{
    Time series of the mean NRMSE improvement for CIWC over a 15-day forecast. The line shows the NRMSE improvement (\%) for the CIWC variable, averaged over all pressure levels, where negative values indicate an improvement.
    }
    \label{fig:ciwc_timeseries}
\end{figure}

\begin{table*}[ht]
\centering
\caption{Ablation study results (average RMSE over the first 5 days of the forecast). Lower is better. Best results are in bold.}
\label{tab:ablation_results}
\resizebox{\textwidth}{!}{%
\begin{tabular}{lccccccccc}
\toprule
& \multicolumn{4}{c}{\textbf{Cloud Microphysics}} & \multicolumn{5}{c}{\textbf{Background Variables}} \\
\cmidrule(lr){2-5} \cmidrule(lr){6-10}
\textbf{Model} & \textbf{CIWC50} & \textbf{CLWC100} & \textbf{CRWC250} & \textbf{CSWC600} & \textbf{U500} & \textbf{V500} & \textbf{Z500} & \textbf{T600} & \textbf{Q600} \\
& ($\times 10^{-8}$ g/kg) & ($\times 10^{-8}$ g/kg) & ($\times 10^{-8}$ g/kg) & ($\times 10^{-5}$ g/kg) & (m/s) & (m/s) & (m$^2$/s$^{2}$) & (K) & ($\times 10^{-4}$ g/kg) \\
\midrule
Baseline & 1.059 & 1.318 & 1.010 & 4.359 & 2.857 & 2.903 & 139.52 & 0.898 & 6.908 \\
w/o (MP, IC) & 1.012 & 0.892 & 0.889 & 4.346 & 2.857 & 2.899 & 138.95 & 0.900 & 6.899 \\
w/o IC & 1.053 & 0.968 & 0.961 & 4.341 & 2.832 & 2.874 & 137.23 & 0.890 & 6.886 \\
\textbf{AviaSafe Model (Ours)} & \textbf{0.956} & \textbf{0.875} & \textbf{0.863} & \textbf{4.340} & \textbf{2.826} & \textbf{2.871} & \textbf{135.81} & \textbf{0.889} & \textbf{6.855} \\
\bottomrule
\end{tabular}%
}
\end{table*}

\noindent\textbf{Long-range Superiority in CIWC Forecasting.}
To further investigate the model's long-range performance on our primary forecast target, Cloud Ice (CIWC), we plot the mean NRMSE improvement percentage for up to 15 days in Figure~\ref{fig:ciwc_timeseries}. This plot clearly illustrates the advantage of AviaSafe.

The performance of both models is very close in the short-range (1-2 days), but the advantage of AviaSafe becomes significant in the 3-7 day range, which is critical for medium-range forecasting. More importantly, this advantage does not decay over time but continues to grow in the long-range forecast. Over the entire 15-day period, the average NRMSE for CIWC shows a consistent reduction, with the majority of improvement concentrated in the latter half of the forecast.

Given that the accurate forecasting of CIWC is crucial for applications like aviation safety, this substantial and sustained advantage in medium and long-range prediction demonstrates the immense potential of our AviaSafe model in these safety-critical domains.



\subsection{Ablation Studies}

To validate the effectiveness of our proposed innovations and the necessity of our hierarchical architecture, we conduct a series of comprehensive ablation studies. We design the following three variants and compare them against the full AviaSafe Mode:
\begin{itemize}
    \item \textbf{w/o (MP, IC):} The decoupled model without the Mask Predictor and its guidance mechanism.
    \item \textbf{w/o IC:} The decoupled model without the IC Module, using only diagnostic cloud masks.
\end{itemize}

\paragraph{Results and Analysis.}
Table~\ref{tab:ablation_results} reports the average RMSE over the first 5 forecast days for nine representative variables. We evaluate four cloud microphysics variables (CIWC, CLWC, CRWC, CSWC) at physically relevant pressure levels and five background variables (Z500, T600, U500, V500, Q600) that summarize the large-scale dynamical, thermodynamical, kinematic, and moisture fields.

We observe a consistent trend toward lower errors as components are added, with the full model performing best overall. First, the w/o (MP, IC) variant outperforms the Baseline on most variables (8/9), indicating that task decoupling alone yields a substantial gain, with a minor degradation on T600.

Second, adding the mask-prediction branch without physics priors (w/o IC) further reduces RMSE on the background variables relative to w/o (MP, IC), but slightly increases errors on the cloud variables. This suggests that unguided masks can inject noisy spatial guidance into the cloud pathway while still benefiting the large-scale fields.

Finally, the full model (AviaSafe), which integrates mask prediction and physics-informed IC features, achieves the best performance across all reported variables. Relative to the next-best variant, it reduces RMSE by 5.5\% on CIWC50 and 2.9\% on CRWC250, and yields additional, albeit modest, improvements on the background variable. These results indicate that physics-informed guidance stabilizes the learned masks and converts them into effective spatial priors for cloud prediction, while also providing auxiliary gradients tied to T and Q that modestly benefit the background fields.

\begin{figure}[ht]
    \centering
    \includegraphics[width=0.8\linewidth]{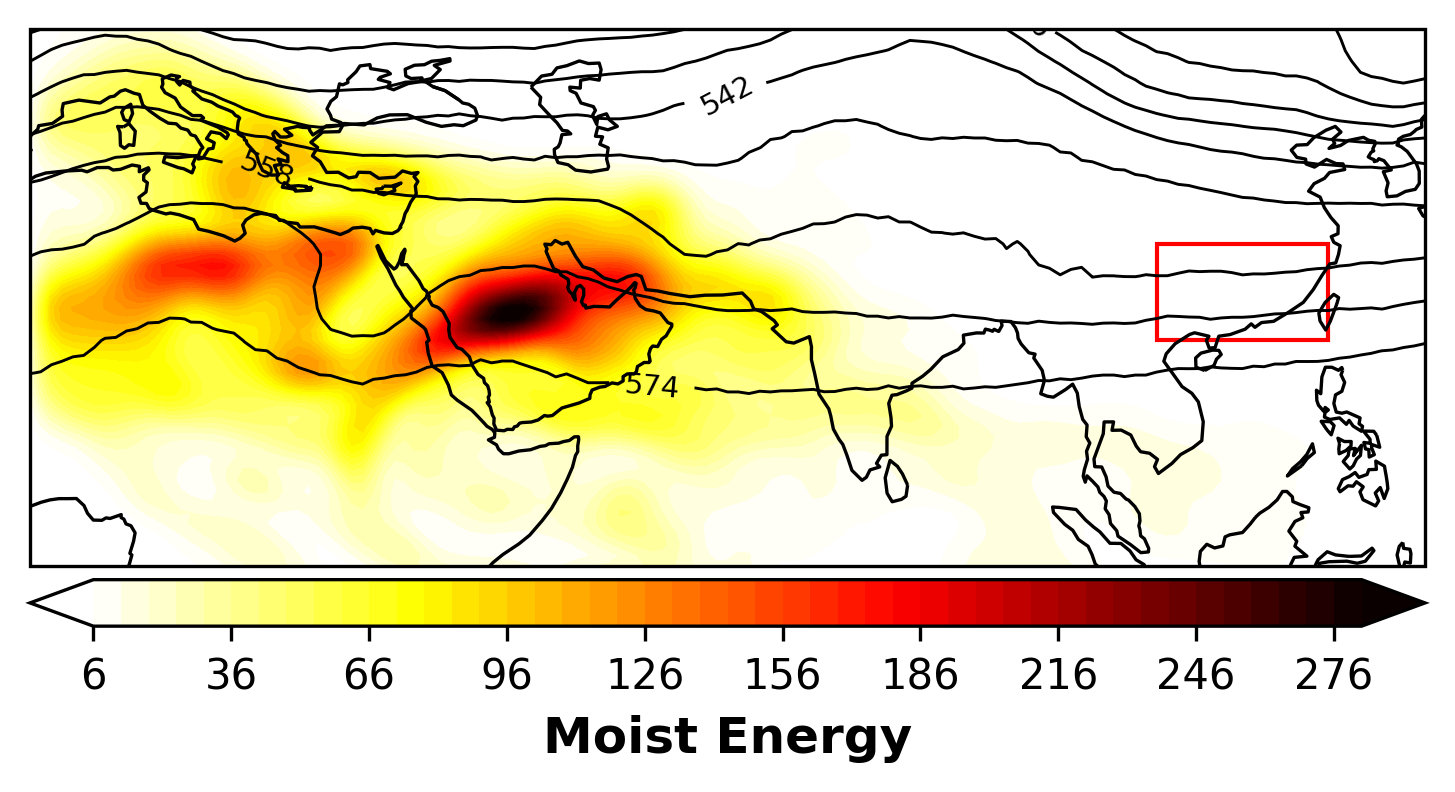}
    \caption{
        CNOP-derived spatial importance map for the moist energy norm. Shading shows the CNOP initial-perturbation pattern; black contours denote the initial 500 hPa geopotential height; the red box marks the target region.
    }
    \label{fig:cnop_energy}
\end{figure}

\begin{figure*}[t]
    \centering
    \includegraphics[width=0.8\linewidth]{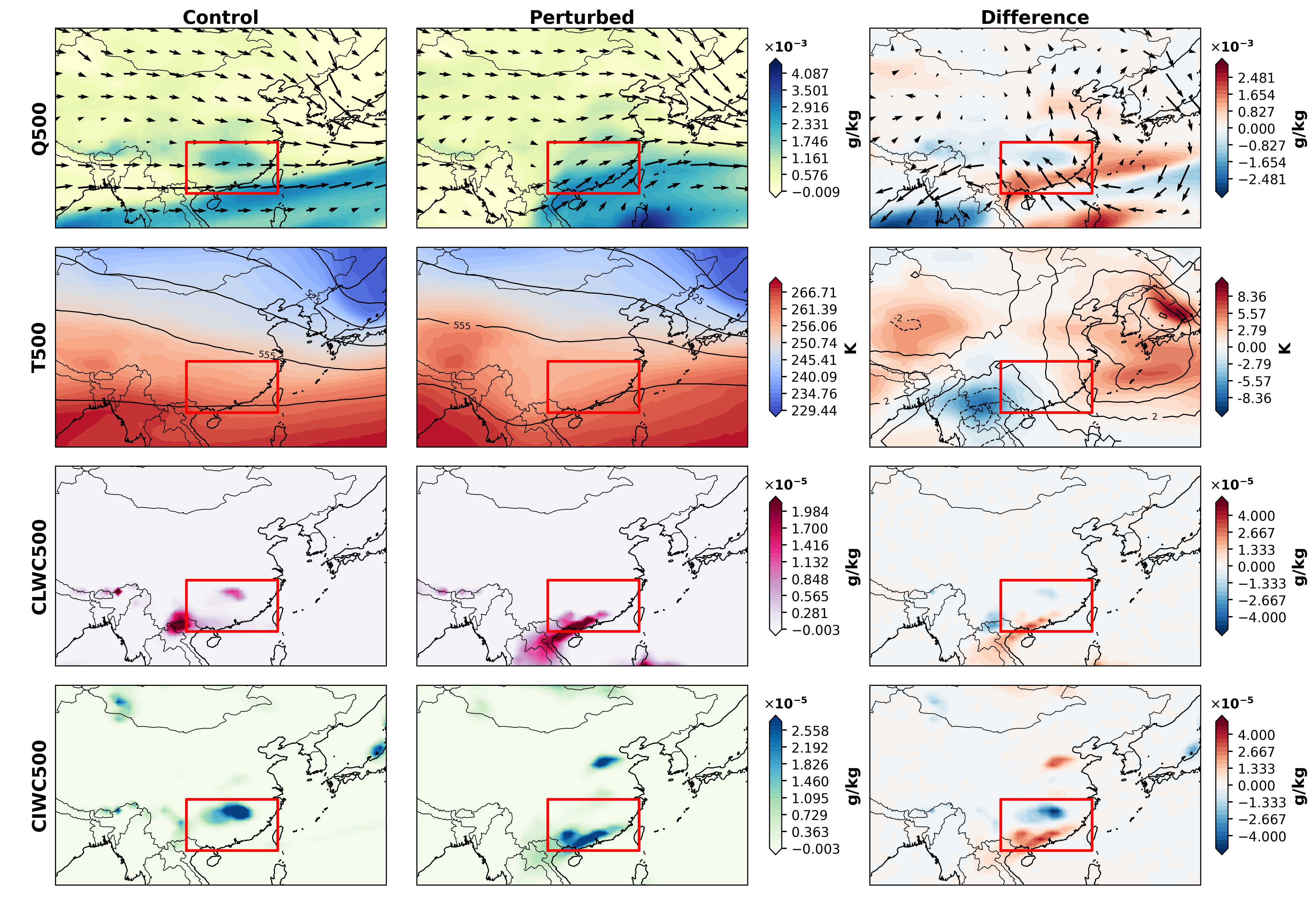}
    \caption{
    Comparison of 72-h forecasts at 500 hPa for CTRL vs CNOP-perturbed initial conditions. 
    Initialization: 2024-01-04 12:00 UTC. 
    The map domain spans 82°E–140°E in longitude and 15°N–50°N in latitude. 
    The red bounding box (105°E–121°E, 21°N–30°N) highlights the target region of interest. 
    Rows (top to bottom): Q, T, CLWC, and CIWC. Columns: control forecast (Control, left), forecast with a CNOP initial perturbation (Perturbed, center), and the Perturbed \- Control difference (right). Units are as indicated by the color bars.
    }
    \label{fig:fcst_err}
\end{figure*}

\subsection{Analysis of Physical Interpretability}
\label{sec:analysis}

To investigate the physical interpretability of our deep learning framework, we select a HIWC clouds formation event observed in January 2024 as a case study. To further bridge data-driven insights with atmospheric dynamical understanding, we adapt the Conditional Nonlinear Optimal Perturbation (CNOP) \cite{npg-10-493-2003} framework from meteorological predictability to spatiotemporal forecasting, in the same spirit as adversarial perturbation analyses~\cite{moosavi2017universal} that reveal critical input regions \cite{qj.2109}. 

Specifically, we formulate a constrained optimization that seeks the initial-state perturbation maximizing the change in the moist energy norm over the target region at a 3-day forecast lead time. Leveraging automatic differentiation, we derive the CNOP solution and obtain a spatial importance map (Figure~\ref{fig:cnop_energy}), conceptually analogous to saliency maps but tailored to atmospheric dynamical systems.

The CNOP analysis identifies a primary importance signal over the Arabian Peninsula (Figure~\ref{fig:cnop_energy}), aligning with an upper-level ridge. As it propagates eastward, this feature transports moisture into the target region, providing favorable conditions for the formation of HIWC clouds.

To validate this physical mechanism, we compare counterfactual forecasts with and without the CNOP initial perturbation (Figure~\ref{fig:fcst_err}). The perturbed simulation amplifies eastward-propagating atmospheric wave patterns, leading to enhanced moisture convergence and downstream cloud microphysical responses. Importantly, the model exhibits phase-dependent behavior: despite ample moisture, ice-phase formation is suppressed where temperatures are unfavorable.

This case study suggests that the architecture responds in a manner consistent with known large-scale dynamics and thermodynamics, translating synoptic patterns into locally coherent cloud microphysical predictions—an important step toward physically grounded interpretability in environmental forecasting.

\section{Conclusion}

AviaSafe is a hierarchical deep learning framework that integrates physics-informed priors via the Icing Condition (IC) index for medium-range, phase-resolved cloud microphysics forecasting at the global scale. To our knowledge, it is the first data-driven model to produce global forecasts of cloud microphysics (CLWC, CIWC, CRWC, CSWC) required for aviation. By combining IC-driven guidance with a mask-informed hierarchical architecture, AviaSafe outperforms our baseline, which is one of the state-of-the-art deep learning methods, on 93.7\% of variable–lead-time pairs and delivers comparable or better skill than operational NWP baselines on key atmospheric variables.
From a practical application perspective, AviaSafe's accurate forecast results provide meaningful value for ensuring the safety of lives and property in aviation.
In subsequent work, we will focus on addressing limitations such as high-resolution, longer-term forecast performance, and the incorporation of observational data to enhance the capability of cloud forecasting for aviation safety.
\section{Acknowledgments}
This work was supported by AI for Science Program, Shanghai Municipal Commission of Economy and Informatization (Grant No.2025-GZL-RGZN-BTBX-02017).
{
    \small
    \bibliographystyle{ieeenat_fullname}
    \bibliography{main}
}

\clearpage
\setcounter{page}{1}
\maketitlesupplementary

\section{Mathematical Formulation and Implementation of CNOP}
\label{sec:cnop_details}

To investigate the physical sensitivity of our model, we compute the Conditional Nonlinear Optimal Perturbation (CNOP). The CNOP represents the initial perturbation $\delta \mathbf{x}_0^*$ that maximizes the forecast error at optimization time $T$ within a target region, subject to an initial energy constraint.

\subsection{Moist Energy Norm}

The magnitude of the perturbation is measured using the moist total energy norm: 
\begin{equation}
\begin{split}
  \|\mathbf{x}'\|^2 = \frac{1}{2} \int_D \left( \right. & u'^2 + v'^2 + \frac{c_p}{T_r} T'^2 \\
  & + R_a T_r \left(\frac{p_s'}{p_r}\right)^2 + \frac{L^2}{c_p T_r} q'^2 \left. \right) dD
\end{split}
\label{eq:moist_energy_norm}
\end{equation}

\noindent where $u', v'$ represent the zonal and meridional wind perturbations, and $T', p_s', q'$ denote perturbations in temperature, surface pressure, and specific humidity, respectively. The constants are defined as: specific heat at constant pressure $c_p = 1004 \, \text{J K}^{-1} \text{kg}^{-1}$, latent heat of condensation $L = 2.5 \times 10^6 \, \text{J kg}^{-1}$, and dry air gas constant $R_a = 287 \, \text{J K}^{-1} \text{kg}^{-1}$. The reference values are set to $T_r = 270 \, \text{K}$ and $p_r = 1000 \, \text{hPa}$. This norm enforces physically meaningful perturbations.

\subsection{Optimization Algorithm}
We employ the Spectral Projected Gradient (SPG) method combined with the Barzilai-Borwein (BB) step size strategy to solve the constrained optimization problem efficiently. 


The optimization process involves the following steps:
\begin{enumerate}
    \item \textbf{Gradient Calculation:} Computing the gradient of the cost function $J$ with respect to the initial state using the adjoint mechanism (via automatic differentiation).
    \item \textbf{Spectral Step Size:} Updating the step size $\alpha_k$ using the BB formula $\alpha_k = \frac{\mathbf{s}^T_{k-1} \mathbf{s}_{k-1}}{\mathbf{s}^T_{k-1} \mathbf{y}_{k-1}}$, where $\mathbf{s}_{k-1}$ is the state difference and $\mathbf{y}_{k-1}$ is the gradient difference from the previous iteration.
    \item \textbf{Non-monotone Line Search:} Performing a line search to ensure global convergence while allowing occasional increases in the objective function (controlled by memory depth $M$) to escape local optima.
    \item \textbf{Projection:} Projecting the updated perturbation onto the constraint ball defined by the moist energy norm $\|\delta \mathbf{x}_0\| \le \xi$.
\end{enumerate}

\begin{algorithm}[h]
\caption{CNOP Calculation via Spectral Projected Gradient (SPG)}
\label{alg:cnop}
\begin{algorithmic}[1]
\Require Pre-trained Model $\mathcal{M}$, Initial State $\mathbf{x}_0$, Constraint $\xi$, Max Iterations $K_{max}$, Memory $M$.
\Ensure Optimal Perturbation $\delta \mathbf{x}^*$.
\State Initialize perturbation $\delta \mathbf{x}_0 \leftarrow \text{random noise}$, projected to $\|\delta \mathbf{x}_0\| \le \xi$.
\State Initialize step size $\alpha_0$, memory $\mathcal{H} = \{-\infty\}$.
\State Compute initial gradient $\mathbf{g}_0 = \nabla J(\delta \mathbf{x}_0)$ via backpropagation.
\State Apply spatial smoothing: $\mathbf{g}_0 \leftarrow \text{Smooth}(\mathbf{g}_0)$.
\For{$k = 0$ to $K_{max}-1$}
    \State \textbf{1. Descent Direction:}
    \State $\quad \mathbf{d}_k = \mathcal{P}_{\xi}(\delta \mathbf{x}_k + \alpha_k \mathbf{g}_k) - \delta \mathbf{x}_k$ \Comment{Projected gradient direction}
    
    \State \textbf{2. Line Search:}
    \State $\quad$ Find step $\lambda \in (0, 1]$ satisfying Armijo condition w.r.t $\max(\mathcal{H})$.
    \State $\quad \delta \mathbf{x}_{k+1} = \delta \mathbf{x}_k + \lambda \mathbf{d}_k$
    
    \State \textbf{3. Update Gradient:}
    \State $\quad$ Compute forecast and loss $J(\delta \mathbf{x}_{k+1})$.
    \State $\quad$ Compute raw gradient $\hat{\mathbf{g}}_{k+1}$.

    \State \textbf{4. Update Step Size:}
    \State $\quad \mathbf{s}_k = \delta \mathbf{x}_{k+1} - \delta \mathbf{x}_k$
    \State $\quad \mathbf{y}_k = \mathbf{g}_{k+1} - \mathbf{g}_k$
    \State $\quad \alpha_{k+1} = \text{clip}\left(\frac{\mathbf{s}_k^T \mathbf{s}_k}{\mathbf{s}_k^T \mathbf{y}_k}, \alpha_{min}, \alpha_{max}\right)$
    
    \State Update memory $\mathcal{H}$ with current loss $J$.
    \State \textbf{if} $\|\mathbf{d}_k\| < \epsilon$ \textbf{then} break
\EndFor
\State \Return $\delta \mathbf{x}^* = \delta \mathbf{x}_{final}$
\end{algorithmic}
\end{algorithm}

\subsection{Hyperparameter Settings}
\label{sec:cnop_params}

To facilitate reproducibility, we detail the specific hyperparameters used in our CNOP experiments. The optimization is performed over a forecast window of $T=72$ hours, targeting the region spanning $21^\circ\text{N}$ to $30^\circ\text{N}$ and $105^\circ\text{E}$ to $121^\circ\text{E}$. 

\vspace{2pt}
\noindent\textbf{Optimization Constraints.} 
The radius of the initial perturbation constraint ball is set to $\xi = 0.6$. The optimization is bounded by a maximum of $K_{max} = 50$ iterations. For the non-monotone line search, we set the memory depth to $M = 10$. The step size $\alpha$ is dynamically updated within the range $[10^{-3}, 50]$.


\begin{figure*}[ht]
    \centering
    \includegraphics[width=1\linewidth]{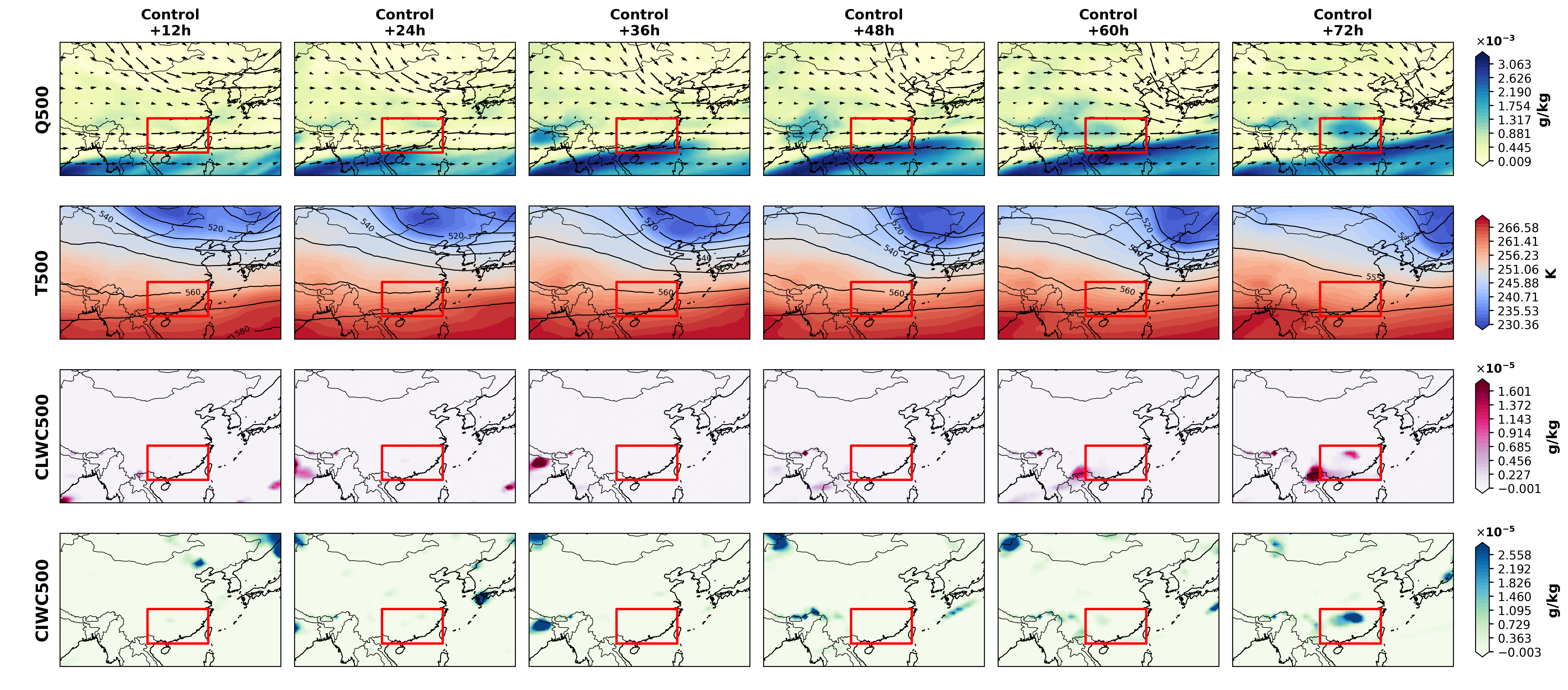}
    \caption{
        \textbf{Spatiotemporal evolution of the control forecast at 500 hPa.} 
        This figure displays the 72-hour forecast trajectory starting from the unperturbed initial state. The columns represent forecast lead times from +12h to +72h. The rows, from top to bottom, illustrate: specific humidity (Q500) overlaid with wind vectors, temperature (T500) with geopotential height contours, cloud liquid water content (CLWC500), and cloud ice water content (CIWC500). The red bounding box highlights the target region for verification. This baseline evolution serves as the reference for assessing the impact of initial state perturbations.
    }
    \label{fig:control_initial_time}
\end{figure*}

\begin{figure*}[ht]
    \centering
    \includegraphics[width=1\linewidth]{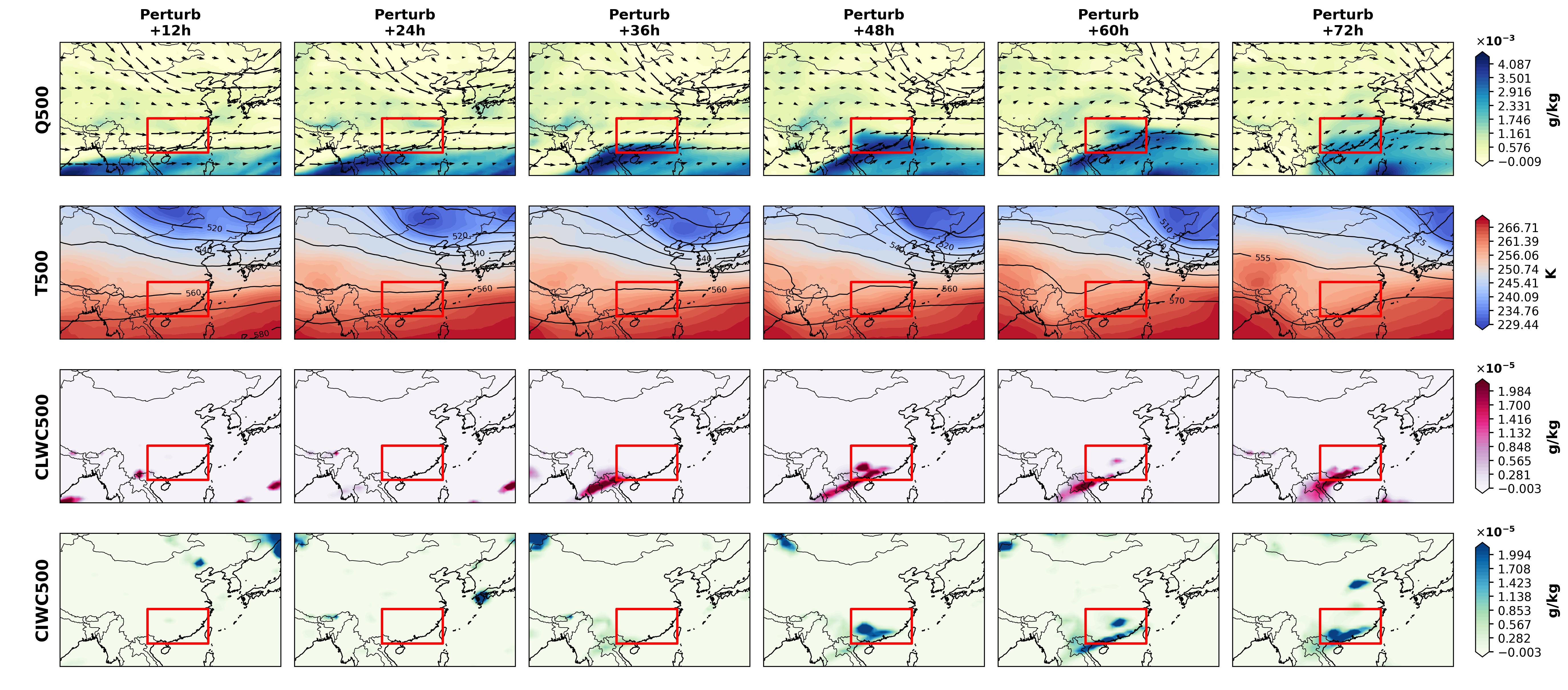}
    \caption{
        \textbf{Spatiotemporal evolution of the CNOP-perturbed forecast at 500 hPa.} 
        This figure shows the forecast results initialized with the CNOP-perturbed state ($\mathbf{x}_0 + \delta \mathbf{x}_0^*$), using the same variable layout and time intervals as the control forecast. 
    }
    \label{fig:perturb_initial_time}
\end{figure*}

\end{document}